\documentclass[english]{gretsi}

\usepackage[utf8]{inputenc}

\usepackage[T1]{fontenc}

\usepackage{amsmath, amssymb, amsfonts}
\usepackage{setspace}
\DeclareMathOperator{\dif}{\mathop{}\!\mathrm{d}\!}

\usepackage{titlesec} 
\titlespacing{\paragraph}{0pt}{\topsep}{1.5ex}

\usepackage{xcolor}

\newcommand\extrafootertext[1]{%
    \bgroup
    \renewcommand\thefootnote{\fnsymbol{footnote}}%
    \renewcommand\thempfootnote{\fnsymbol{mpfootnote}}%
    \footnotetext[0]{#1}%
    \egroup
}
\begin{document}


\titre{A multilevel approach to accelerate the training of Transformers}

\auteurs{
  \auteur{Guillaume}{Lauga}{guillaume.lauga+gretsi@ens-lyon.fr}{1}
  \auteur{Maël}{Chaumette\dag}{mael.chaumettes+gretsi@ens-lyon.fr}{1}
  \auteur{Edgar}{Desainte-Maréville\dag}{edgar.desainte-mareville+gretsi@ens-lyon.fr}{1}
  \auteur{Étienne}{Lasalle\dag}{etienne.lasalle+gretsi@ens-lyon.fr}{1}
  \auteur{Arthur}{Lebeurrier\dag}{arthur.lebeurrier+gretsi@ens-lyon.fr}{1}
}

\affils{
  \affil{1}{ENS de Lyon, CNRS, Inria, Université Claude Bernard Lyon 1, LIP, UMR 5668, 69342, Lyon cedex 07, France
  }
}


\resume{Dans cet article, nous étudions le potentiel des approches multiniveaux pour accélérer l'entraînement des Transformers. En utilisant une interprétation de ces architectures sous forme d'équations différentielles ordinaires (EDO), nous proposons une manière appropriée de varier la discrétisation de l'EDO décrivant ces transformers afin d'accélérer l'apprentissage. Nous validons le potentiel de notre approche en la comparant à un apprentissage standard.}

\abstract{In this article, we investigate the potential of multilevel approaches to accelerate the training of transformer architectures. Using an ordinary differential equation (ODE) interpretation of these architectures, we propose an appropriate way of varying the discretization of these ODE Transformers in order to accelerate the training. We validate our approach experimentally by a comparison with the standard training procedure.}

\maketitle


\section{Introduction}\extrafootertext{\dag : Equal contributions. We  acknowledge the support of the Centre Blaise Pascal's IT test platform at ENS de Lyon (Lyon, France) for its computing facilities. The platform operates the SIDUS solution \cite{quemener2013} developed by Emmanuel Quemener.}
Transformer architectures have become ubiquitous since their introduction \cite{vaswani_attention_2017}. They are state-of-the-art on a large class of problems including image recognition \cite{dosovitskiy2020image, liu2021swin}, speech processing \cite{chang2020end, lu2020exploring} and large language models \cite{radford2019language, brown2020language, touvron2023llama}. 
The performance of these models, which increases with their dimension, is balanced by training and inference costs.
Hence, it is crucial to propose new training methods to better handle their cost. In this paper, we propose to accelerate the training of transformers models by relying on an ODE interpretation of transformer-decoder architectures and multilevel optimization, in a similar fashion as in \cite{kopanicakova_globally_2022,gaedke-merzhauser_multilevel_2021} for ResNet models.

Transformer-decoder networks are composed of an input layer that transforms tokens into embeddings and passes these embeddings to a sequence of transformer blocks which are attention layers followed by feed-forward networks, intertwined with normalization layers. The output is then transformed from embeddings to tokens. 
Similar to previous deep neural architectures, transformers become progressively harder to train as their depth increases, primarily due to large-scale optimization and dependencies on residual connections that makes training unstable, since it amplifies small parameter perturbations \cite{liu2020understanding}.  Reducing the number of parameters is a direct remedy to this problem, at the expense of the network's expressivity. Therefore it becomes more and more interesting to find training approaches able to manage the growth in the number of parameters without reducing performance.

A possible framework to manage this growth is derived from an ODE interpretation of transformer networks \cite{li_ode_2022,baier2020n}. The success of the ODE interpretation of ResNet networks \cite{chen_neural_2019, avelin2021neural} motivated the community to extend such study to other residual networks such as transformer-decoder architectures \cite{li_ode_2022,baier2020n}, at the heart of modern large language models \cite{radford2019language,brown2020language}. 

Our approach is inspired by the work of \cite{kopanicakova_globally_2022,gaedke-merzhauser_multilevel_2021}, whose authors investigate the impact of varying depths of ResNet networks based on the ODE formalism, and train a deep ResNet in a multilevel fashion. 
In order to accelerate the optimization, multilevel algorithms exploit a hierarchy of approximations of the objective function. 
Reducing the problem dimension can lead to significant gains in convergence speed \cite{lauga_iml_2023,lauga_multilevel_2024}. 

\begin{figure}
    \centering
    \includegraphics[trim={1em 1em 0em 0em},clip,width=0.35\textwidth, height=0.35\textwidth]{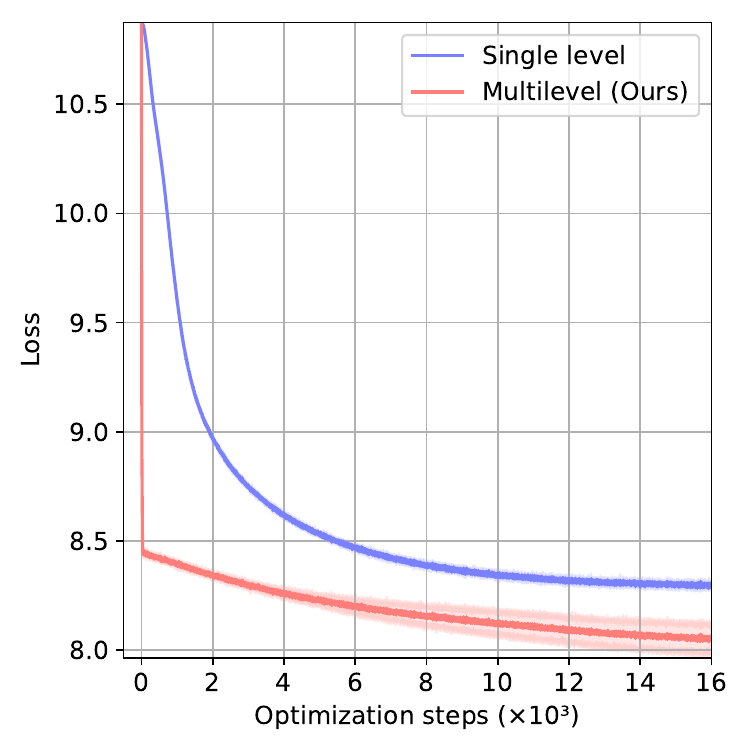} \vspace{-1em}
    \caption{Training loss of the single level algorithm (standard method) in blue and of the multilevel algorithm (our proposed approach) in red with respect to the number of optimization steps (at fine level). The curves are averaged over $6$ seeds. In lighter colors we display the standard deviation of the $6$ training runs.}
    \label{fig:training_loss_steps}
\end{figure}

\paragraph{Related works.} 
A lot of effort has been dedicated to accelerating the training of transformers. A common idea is to increase the size of the network during training, starting from a smaller network. We can divide these approaches depending on the proposed growth method: depth \cite{gong2019efficient}, width  \cite{gu2020transformer} or both \cite{chen2015net2net,chen2021bert2bert,wang2023learning,ding2023network}. 

In \cite{gong2019efficient}, the authors propose a method to transfer knowledge from a shallow model to a deeper one by progressively stacking layers and doubling its depth. This strategy is motivated by observing that lower and upper layer representations exhibit similar distributions. A similar perspective is adopted in \cite{gu2020transformer}, where the authors stack weights and double the width of the model, ensuring that the output remains preserved after the expansion.

In \cite{chen2015net2net,chen2021bert2bert, ding2023network}, authors propose techniques for transferring information stored in a neural network into another neural network with increased width or depth. 
For instance, the Net2DeeperNet method in \cite{chen2015net2net} transforms one layer into two layers by initializing the added layer's weights as the identity. In \cite{ding2023network}, the small network is initialized as a pruned dense network, that gradually extends into the dense network.
Authors in \cite{wang2023learning} also propose a method to simultaneously extend the width and the depth. Their algorithm learns an expansion matrix from a given dataset. All these methods change the size of the network during training, and are therefore not comparable to our proposed approach.

In an adjacent direction, authors of \cite{yang2021speeding} propose to speed up the training by sharing weights across all transformer blocks up until a certain point in the training. This procedure allows the blocks to quickly learn common shared features, therefore bringing the weights closer to the optimal solution faster.

The idea of training neural networks using multilevel techniques has been investigated in several papers \cite{gaedke-merzhauser_multilevel_2021,kopanicakova_globally_2022,zou_multi-level_2024,gratton2024block}. Authors of \cite{gaedke-merzhauser_multilevel_2021,kopanicakova_globally_2022} start from the ODE interpretation of ResNets to create a hierarchy of smaller networks then used in a multilevel fashion by alternating the training of the smaller networks and of the target network, achieving great acceleration. In \cite{gratton2024block}, the context of PDE solving allows the network to be divided into blocks that can be trained separately. Each block targeting different frequency components of the solution, the training is accelerated. 

Lastly, a multilevel approach reducing width and depth of transformer network during training was recently proposed in \cite{zou_multi-level_2024}, saving up to $20\%$ of the total computational cost. The authors introduce a construction of operators that allow to go from one network to the other, reducing (resp. increasing) the depth and the width sequentially.

\paragraph{Contributions.} In this paper, we propose a multilevel approach to train an ODE transformer neural network for sequence generation. By varying the depth, we obtain smaller networks that are easier to train, and we propagate these smaller networks to the fine network, thus accelerating its training. In contrast to \cite{zou_multi-level_2024}, we do not use operators to reduce the width and the depth of the network. We rely on the ODE interpretation to reduce the dimension without any operators.

\paragraph{Outline.} In Section \ref{sec:odeInterpretation}, we present the ODE formalization of transformer networks, and we then introduce our multilevel algorithm for the training. In Section \ref{sec:numericals}, we compare our algorithm to a single level training on a sequence generating task. Finally, we discuss our findings and potential improvements.

\begin{figure}
    \centering
    \includegraphics[width=0.35\textwidth]{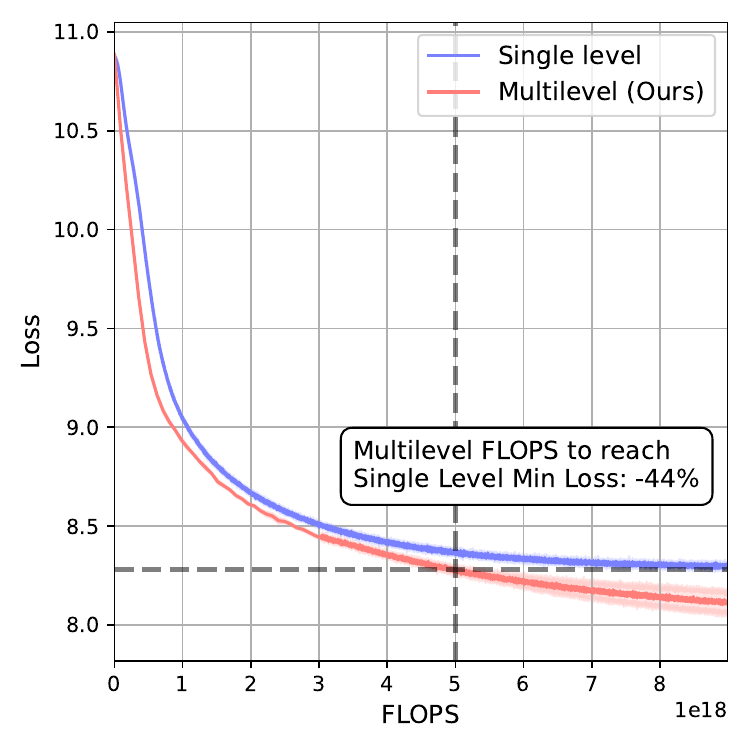}\vspace{-1.5em}
    \caption{Training loss of the single level algorithm (standard method) in blue and of the multilevel algorithm (our proposed approach) in red with respect to FLOPS. The curves are averaged over $6$ seeds. In lighter colors we display the standard deviation of the $6$ training runs.}
    \label{fig:training_flops}
\end{figure}
\section{Multilevel training for ODE transformers}\label{sec:odeInterpretation}

In this section, we present the interpretation of transformer-decoder networks using an ODE \cite{li_ode_2022}. The presentation is slightly different than in \cite{li_ode_2022}, but principles remain similar. This interpretation will then allow us to derive a multilevel algorithm to train transformer networks.

\subsection{ODE Transformers.} Like ResNets, transformer networks use residual connections. The ODE interpretation is well known for ResNets \cite{chen_neural_2019, avelin2021neural}, and is a consequence of these residual connections.

A transformer layer is composed of a self-attention (SA) block followed by a feed-forward (FF) network. The input of these blocks is passed through a Layer Norm (LN) operation, and a residual connection. Denote by $\mathbf x_t$ the input of a transformer layer at time $t$. The output $\mathbf x_{t+1}$ is given by:
\begin{align}
    \mathbf x_{t+1} & = \mathbf x_t + \mathrm{SA}_t(\mathrm{LN}_t(\mathbf x_t)) + \mathrm{FF}_t(\mathrm{LN}_t(\mathbf x_t + \mathrm{SA}_t(\mathrm{LN}_t(\mathbf x_t)))) \nonumber \\
    & = \mathbf x_t + \mathrm{F}(\mathbf x_t,\theta_t)
\end{align}
where $\mathrm{F}(\cdot,\cdot) = \mathrm{SA}_t(\mathrm{LN}_t(\cdot)) + \mathrm{FF}_t(\mathrm{LN}_t(\cdot + \mathrm{SA}_t(\mathrm{LN}_t(\cdot))))$. At time $t$, $\mathrm{F}$ will depend on parameters $\theta_t$. This equation can be seen as the Euler discretization of the following ODE
\begin{equation}
    \frac{\dif \mathbf x(t)}{\dif t} = \mathrm{F}(\mathbf x(t), \theta(t))
\end{equation}
with a discretization step $\Delta t = 1$. Our idea is to use different levels of discretization of this ODE to define a hierarchy of networks. The fine level network will have the desired $N$ (even) number of transformer layers $\mathrm{F}_i$ with parameters $\theta_i$, for $i=1,\ldots,N$. Coarser networks will use less and less layers corresponding to their discretization level.  

\subsection{Multilevel training of transformers.} For the clarity of the presentation, we limit ourselves to two levels. We will denote by $h$ elements of the fine level network, and by $H$ elements of the coarse level network.

In the following, coarse and fine level networks will share input and output layers that respectively transform tokens into embeddings and embeddings into tokens. It is standard practice for input and output layers to share the same weights \cite{radford2019language,brown2020language}. 

\begin{figure}[htb]
    \begin{center}
      \includegraphics[trim={1em 15em 1em 10em},clip,width=0.9\columnwidth]{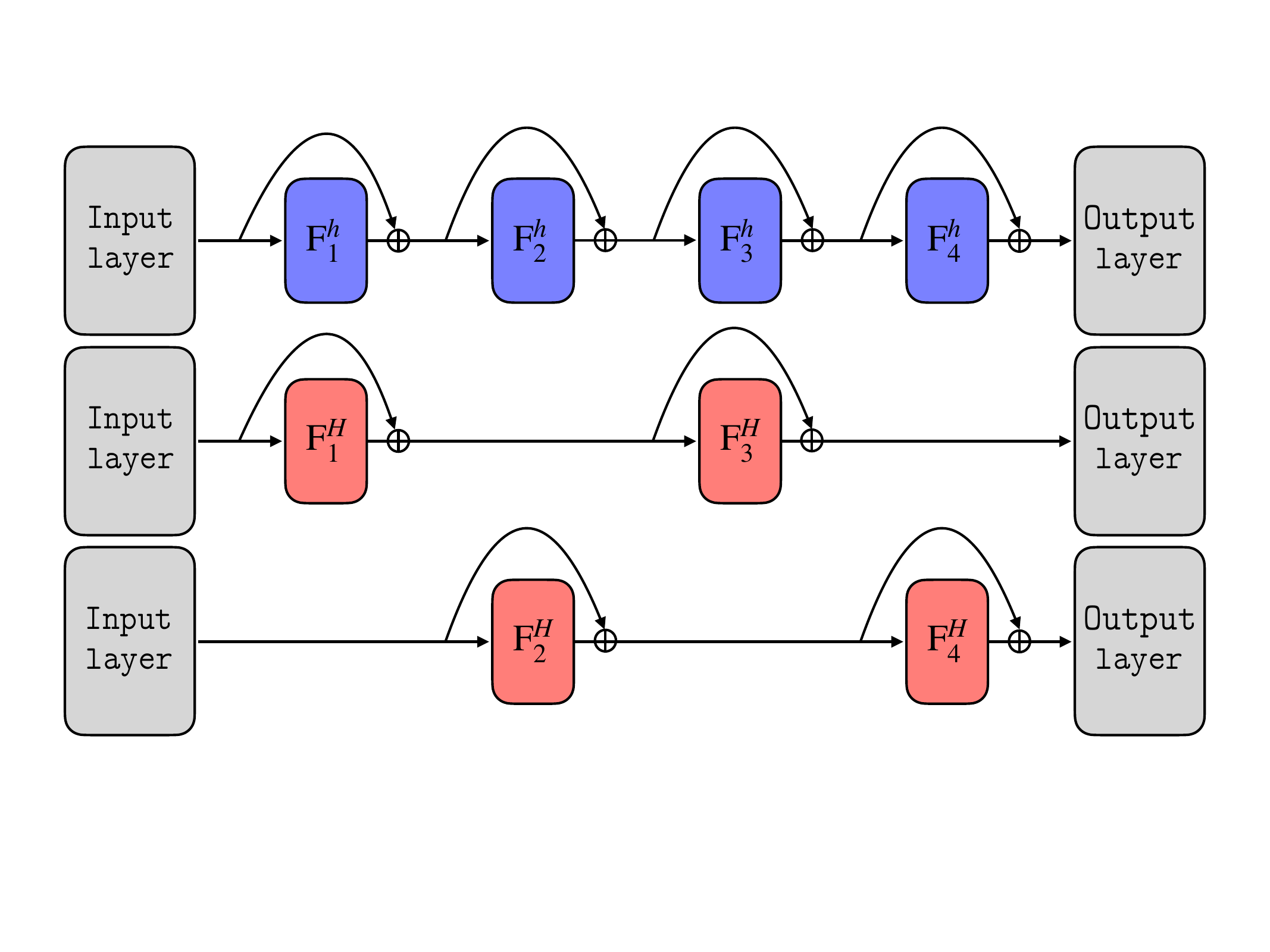}
    \end{center} \vspace{-2em}
    \caption{Scheme of our proposed approach on a network with $4$ transformer layers. The fine level network (\textcolor{blue}{blue} blocks) is decomposed into two coarser networks (\textcolor{red}{red} blocks) that contains even-indexed (resp. odd-indexed) layers. Input and output layers (\textcolor{gray}{gray} blocks) are shared across all networks.}
    \label{fig:scheme}
  \end{figure}

\paragraph{Construction of coarse models.} To coarsen our fine level network, we will divide by $2$ the number of transformer layers. The coarse model has therefore $N/2$ layers $F^H_i$ with parameters $\theta^H_i$ for $i=1,\ldots,N/2$. These layers are the layers of the fine level network, linked by the following relationship
\begin{equation}
    \forall i \in \{1,\ldots,N/2\}, \quad \theta_i^H = \theta_{2i}^h.
\end{equation}
At any point during the training, this relationship holds (without additional memory footprint). After $K$ steps of coarse optimization, we obtain a new set of parameters $(\Tilde{\theta}_i^H)_{1 \leq i \leq N/2}$. We then update fine level parameters by prolongating the coarse parameters. For all $i\in \{1,\ldots,N\}$
\begin{align}
    \Tilde{\theta}_{2i}^h & = \Tilde{\theta}_i^H, \nonumber \\
    \Tilde{\theta}_{2i+1}^h & = (1-\delta) \theta_{2i+1}^h + \delta \Tilde{\theta}_i^H, \label{eq:prolongation}
\end{align}
where $\delta \in [0,1]$ is an averaging constant. Here we follow the example of \cite{zou_multi-level_2024}. The simplest setting would be to set $\delta =1$, and would be substantiated by the literature on numerical solution of ODEs (see \cite{kopanicakova_globally_2022,gaedke-merzhauser_multilevel_2021} and references therein). However, due to the low number of layers in our context (a dozen versus a few thousands in \cite{kopanicakova_globally_2022,gaedke-merzhauser_multilevel_2021}), we would lose too much information at the fine level. Moreover, one can see that only training even layers leads the training to be quite asymmetrical. Therefore we propose to \textit{train two coarse models}. The second coarse model is linked to the fine model by 
\begin{equation}
    \forall i \in \{1,\ldots,N/2\}, \quad \theta_i^H = \theta_{2i-1}^h,
\end{equation}
with the symmetric of Equation \ref{eq:prolongation} to prolongate the coarse parameters to the fine levels. We display in Figure \ref{fig:scheme} an example of a fine level network with $4$ layers and its two coarse models.
One training step on a coarse model costs less than an training step on the fine model and training a smaller network is easier. Hence, the loss of expressivity by using smaller networks should be outweighed by the computational and training gains.

\paragraph{Optimization parameters at coarse level.} It remains now to address the training of the coarse models. Both coarse models see the same data points as the fine level, in the same order. We tried to vary the data points and the order, without any noticeable effect. The number of tokens seen at each optimization step also remains the same. In a few words, we train each coarse model like the fine model. After training both coarse models, we always compute one optimization step for the fine model.

A notable difference with standard multilevel approaches is that we do not share gradient information between levels. Standard approaches impose coherence between levels using this information \cite{lauga_iml_2023}. Due to the stochastic nature of the gradient, in our context, adopting this coherence leads to be coherent with noisy information.
\vspace{-0.5em}
\section{Numerical experiments} \label{sec:numericals}
In this section, we numerically compare the training speed of our approach against the standard single level approach.

\paragraph{Dataset.} We use a portion of the FineWeb-Edu dataset: a billion of GPT-2 tokens, sampled from the 10 billions smaller version \cite{penedo2024finewebdatasetsdecantingweb}.
This dataset contains educational pages of the FineWeb dataset.

\paragraph{Architecture choice.} For our experiments we consider accelerating the training of a transformer-decoder architecture that takes as input sequences $256$ tokens obtained with the GPT-2 tokenizer\footnote{\href{https://github.com/openai/tiktoken}{tiktoken} used with \texttt{tiktoken.get\_encoding("gpt-2")}} in a vocabulary of size $50257$.
The input layer embeds the tokens into a space of dimension $256$, after which follows $12$ transformer blocks with $8$ attention heads each.
The output layer then reverts the embedding operation. 
Across all levels, the input and output layers remain untouched, and we only reduce the number of transformer blocks to define our coarse models. The number of parameters of this architecture is $22,368,512$, which allows training the fine model and the coarse models on a single GPU with $16$ GB of RAM.

\paragraph{Hyperparameters for training.} Each model was trained with Stochastic Gradient Descent (SGD). This choice is not the state-of-the-art for training transformer models for sequence generation, but is easier to handle when dealing with the training of several related networks (see the discussion at the end). $16000$ steps are computed with a batch size of $32$ sequences of length $256$, accumulated to reach a total batch size of $262144$ tokens.
Therefore, the network is trained for a little bit more than $4$ epochs.
All training runs follow a learning rate schedule with $715$ steps of linear warm-up followed by a cosine decay to zero. Minimum and maximum learning rates were set to $1.2 \times 10^{-4}$ and $1.2 \times 10^{-3}$ respectively \cite{radford2019language,brown2020language}.
Hyperparameters for the coarse models are identical, except for the learning rate that remains constant equal to $1.2 \times 10^{-3}$.

\paragraph{Multilevel hyperparameters.} We use two coarse models: one that trains odd-indexed transformers blocks (1,3,5,7,9,11), the other that trains even-indexed (2,4,6,8,10,12) transformer blocks. Each coarse model has $17,649,920$ parameters. After one step of fine level optimization, we use the coarse models for $35$ steps. During these steps, each coarse model is trained for $100$ steps, before propagating their weights to the fine level network. Both coarse networks are trained using the same optimizer as the fine network. For both coarse models, we set the averaging constant $\delta$ to $0.25$ following \cite{zou_multi-level_2024} and our own experiments. After these $35$ steps, we only train the fine level network. In practice, when using coarse levels long after these starting steps, we observed diminishing returns with respect to the computational cost, therefore we only used them at the start. 

\paragraph{Comparison metrics.} In order to show that a multilevel approach can accelerate the training, we compare the training losses with respect to the iterations of our approach and of the single level approach.
Such comparison is at our advantage since one iteration of our algorithm encompasses several coarse iterations.
Thus, to be fair to the single level algorithm we also compare the training loss with respect to the number of Floating Point Operations (FLOP)s computed per iteration.
We estimate the number of FLOPs of one iteration by estimating the number of FLOPs required to do one forward pass of the networks.
Then we follow similar works that estimate that one training step requires in FLOPs the equivalent of 3 forward pass (cf this \href{https://epoch.ai/blog/backward-forward-FLOP-ratio}{report}\footnote{https://epoch.ai/blog/backward-forward-FLOP-ratio} or \cite{kaplan2020scaling}).

\paragraph{Results.}
Training curves are averaged over $6$ training runs using $6$ different seeds. We display in  Figure \ref{fig:training_loss_steps} the training loss of our multilevel approach and of the single level approach with respect to the number of optimization steps. In Figure \ref{fig:training_flops}, we display losses with respect to the number of FLOPs. Our algorithm achieves the same loss as the single level training in $16000$ steps, while reducing the total number of FLOPs by $44\%$, demonstrating the efficiency of our multilevel method in accelerating the training of transformer networks.

\paragraph{Discussion.} As mentioned earlier, \cite{zou_multi-level_2024} performs multilevel on a transformer-decoder. In order to compare to their method, we need to train larger architectures such as GPT-2 \cite{radford2019language}, as they did. Due to the higher number of hyperparameters requiring tuning for their method to work (compared to ours), we could not afford to do this expensive tuning yet. Hence, we leave a comparison for later works.

However, their coarse level is obtained by reducing the width of each layer and the depth of the network while our multilevel approach focuses on the depth of a transformer-decoder as this method respects the ODE interpretation detailed in Section \ref{sec:odeInterpretation}.
We are aware that SGD is not the preferred algorithm for language tasks \cite{zhao2024deconstructing}. Showing that our approach can accelerate the training with the vanilla algorithm is a first but necessary step towards the acceleration of more complex algorithms. 
Notably, one question needs to be solved in order to be competitive with Adam or AdamW: the interaction between the momentum at fine and coarse level.
In a convex setting, this momentum does not seem to be a problem \cite{lauga_iml_2023}, but in deep learning settings it is \cite{gratton2024block}. To the best of our knowledge, no multilevel algorithm handle it in a satisfying manner.
The question is either omitted \cite{zou_multi-level_2024,gaedke-merzhauser_multilevel_2021,kopanicakova_globally_2022} or circumvented using restarting techniques \cite{gratton2024block}.

\section{Conclusion}
In this article we proposed a multilevel approach to accelerate the training of transformer networks. Our approach is based on an ODE interpretation of these networks, which allows us to vary their size by varying the discretization that solves the associated ODE. We obtained impressive savings in FLOPs with respect to the single level training. At the moment, the setting of these experiments is limited. It calls for more experiments, and also more theoretical developments to better understand interactions between the training of finer and coarser networks, notably the momentum. We also need to compare the robustness of the trained network with respect to test tasks to completely validate the benefit of our approach.


{ \footnotesize
\begin{spacing}{0.8}
\bibliography{references}

\begin{thebibliography}{10}

\bibitem{avelin2021neural}
B.~Avelin and K.~Nystr{\"o}m.
\newblock Neural {ODE}s as the deep limit of {ResNets} with constant weights.
\newblock {\em Analysis and Applications}, 19(03):397--437, 2021.

\bibitem{baier2020n}
A.~Baier-Reinio and H.~De~Sterck.
\newblock N-ode transformer: A depth-adaptive variant of the transformer using
  neural ordinary differential equations.
\newblock {\em Preprint arXiv:2010.11358}, 2020.

\bibitem{brown2020language}
T.~Brown, B.~Mann, N.~Ryder, M.~Subbiah, J.~D. Kaplan, P.~Dhariwal,
  A.~Neelakantan, P.~Shyam, G.~Sastry, A.~Askell, et~al.
\newblock Language models are few-shot learners.
\newblock {\em NeurIPS}, 33:1877--1901, 2020.

\bibitem{chang2020end}
X.~Chang, W.~Zhang, Y.~Qian, J.~Le~Roux, and S.~Watanabe.
\newblock End-to-end multi-speaker speech recognition with transformer.
\newblock In {\em ICASSP 2020}, pages 6134--6138. IEEE, 2020.

\bibitem{chen2021bert2bert}
C.~Chen, Y.~Yin, L.~Shang, X.~Jiang, Y.~Qin, F.~Wang, Z.~Wang, X.~Chen, Z.~Liu,
  and Q.~Liu.
\newblock bert2bert: Towards reusable pretrained language models.
\newblock {\em Preprint arXiv:2110.07143}, 2021.

\bibitem{chen_neural_2019}
R.~T.~Q. Chen, Y.~Rubanova, J.~Bettencourt, and D.~Duvenaud.
\newblock Neural {Ordinary} {Differential} {Equations}, December 2019.
\newblock arXiv:1806.07366.

\bibitem{chen2015net2net}
T.~Chen, I.~Goodfellow, and J.~Shlens.
\newblock Net2net: Accelerating learning via knowledge transfer.
\newblock {\em Preprint arXiv:1511.05641}, 2015.

\bibitem{ding2023network}
N.~Ding, Y.~Tang, K.~Han, C.~Xu, and Y.~Wang.
\newblock Network expansion for practical training acceleration.
\newblock In {\em IEEE/CVF CVPR}, pages 20269--20279, 2023.

\bibitem{dosovitskiy2020image}
A.~Dosovitskiy, L.~Beyer, A.~Kolesnikov, D.~Weissenborn, X.~Zhai,
  T.~Unterthiner, M.~Dehghani, M.~Minderer, G.~Heigold, S.~Gelly, et~al.
\newblock An image is worth 16x16 words: {Transformers} for image recognition
  at scale.
\newblock {\em Preprint arXiv:2010.11929}, 2020.

\bibitem{gaedke-merzhauser_multilevel_2021}
L.~Gaedke-Merzhäuser, A.~Kopaničáková, and R.~Krause.
\newblock Multilevel minimization for deep residual networks.
\newblock {\em ESAIM: Proceedings and Surveys}, 71:131--144, August 2021.

\bibitem{gong2019efficient}
L.~Gong, D.~He, Z.~Li, T.~Qin, L.~Wang, and T.~Liu.
\newblock Efficient training of bert by progressively stacking.
\newblock In {\em ICML}, pages 2337--2346. PMLR, 2019.

\bibitem{gratton2024block}
S.~Gratton, V.~Mercier, E.~Riccietti, and P.~L. Toint.
\newblock A block-coordinate approach of multi-level optimization with an
  application to physics-informed neural networks.
\newblock {\em Computational Optimization and Applications}, 89(2):385--417,
  2024.

\bibitem{gu2020transformer}
X.~Gu, L.~Liu, H.~Yu, J.~Li, C.~Chen, and J.~Han.
\newblock On the transformer growth for progressive bert training.
\newblock {\em Preprint arXiv:2010.12562}, 2020.

\bibitem{kaplan2020scaling}
J.~Kaplan, S.~McCandlish, T.~Henighan, T.~B. Brown, B.~Chess, R.~Child,
  S.~Gray, A.~Radford, J.~Wu, and D.~Amodei.
\newblock {Scaling Laws for Neural Language Models}.
\newblock {\em CoRR}, abs/2001.08361, 2020.

\bibitem{kopanicakova_globally_2022}
A.~Kopani{\v c}{\'a}kov{\'a} and R.~Krause.
\newblock Globally {Convergent} {Multilevel} {Training} of {Deep} {Residual}
  {Networks}, June 2022.

\bibitem{lauga_multilevel_2024}
G.~Lauga.
\newblock {\em Multilevel proximal methods and application to image
  restoration}.
\newblock phdthesis, {\'E}cole Normale Sup{\'e}rieure de Lyon, December 2024.

\bibitem{lauga_iml_2023}
G.~Lauga, E.~Riccietti, N.~Pustelnik, and P.~Gon{\c c}alves.
\newblock {IML} {FISTA}: {A} {Multilevel} {Framework} for {Inexact} and
  {Inertial} {Forward}-{Backward}. {Application} to {Image} {Restoration}.
\newblock {\em SIAM Journal on Imaging Sciences}, 17(3):1347--1376, 2024.

\bibitem{li_ode_2022}
B.~Li, Q.~Du, T.~Zhou, Y.~Jing, S.~Zhou, X.~Zeng, T.~Xiao, J.~Zhu, X.~Liu, and
  M.~Zhang.
\newblock {ODE} {Transformer}: {An} {Ordinary} {Differential}
  {Equation}-{Inspired} {Model} for {Sequence} {Generation}, March 2022.
\newblock arXiv:2203.09176.

\bibitem{liu2020understanding}
L.~Liu, X.~Liu, J.~Gao, W.~Chen, and J.~Han.
\newblock Understanding the difficulty of training transformers.
\newblock {\em Preprint arXiv:2004.08249}, 2020.

\bibitem{liu2021swin}
Z.~Liu, Y.~Lin, Y.~Cao, H.~Hu, Y.~Wei, Z.~Zhang, S.~Lin, and B.~Guo.
\newblock Swin transformer: {Hierarchical} vision transformer using shifted
  windows.
\newblock In {\em IEEE/CVF ICCV}, pages 10012--10022, 2021.

\bibitem{lu2020exploring}
L.~Lu, C.~Liu, J.~Li, and Y.~Gong.
\newblock Exploring transformers for large-scale speech recognition.
\newblock {\em Preprint arXiv:2005.09684}, 2020.

\bibitem{penedo2024finewebdatasetsdecantingweb}
G.~Penedo, H.~Kydlíček, L.~Ben allal, A.~Lozhkov, M.~Mitchell, C.~Raffel,
  L.~Von Werra, and T.~Wolf.
\newblock The {FineWeb Datasets: Decanting the Web for the Finest Text Data at
  Scale}, 2024.

\bibitem{quemener2013}
E.~Quemener and M.~Corvellec.
\newblock {SIDUS}—the {Solution} for {Extreme} {Deduplication} of an
  {Operating} {System}.
\newblock {\em Linux Journal}, 2013.

\bibitem{radford2019language}
A.~Radford, J.~Wu, R.~Child, D.~Luan, D.~Amodei, I.~Sutskever, et~al.
\newblock Language models are unsupervised multitask learners.
\newblock {\em OpenAI blog}, 1(8):9, 2019.

\bibitem{touvron2023llama}
H.~Touvron, T.~Lavril, G.~Izacard, X.~Martinet, M.A. Lachaux, et~al.
\newblock Llama: {Open} and efficient foundation language models.
\newblock {\em Preprint arXiv:2302.13971}, 2023.

\bibitem{vaswani_attention_2017}
A.~Vaswani, N.~Shazeer, N.~Parmar, J.~Uszkoreit, L.~Jones, A.~N. Gomez, Ł.
  Kaiser, and I.~Polosukhin.
\newblock Attention is {All} you {Need}.
\newblock 2017.

\bibitem{wang2023learning}
P.~Wang, R.~Panda, L.~T. Hennigen, P.~Greengard, L.~Karlinsky, R.~Feris, D.~D.
  Cox, Z.~Wang, and Y.~Kim.
\newblock Learning to grow pretrained models for efficient transformer
  training.
\newblock {\em Preprint arXiv:2303.00980}, 2023.

\bibitem{yang2021speeding}
S.~Yang, L.~Hou, X.~Song, Q.~Liu, and D.~Zhou.
\newblock Speeding up deep model training by sharing weights and then
  unsharing.
\newblock {\em Preprint arXiv:2110.03848}, 2021.

\bibitem{zhao2024deconstructing}
R.~Zhao, D.~Morwani, D.~Brandfonbrener, N.~Vyas, and S.~Kakade.
\newblock Deconstructing what makes a good optimizer for language models.
\newblock {\em Preprint arXiv:2407.07972}, 2024.

\bibitem{zou_multi-level_2024}
L.~Zou, H.~Zhang, and Y.~Deng.
\newblock A {Multi}-{Level} {Framework} for {Accelerating} {Training}
  {Transformer} {Models}, April 2024.
\newblock arXiv:2404.07999.

\end{thebibliography}
\end{spacing}
}


\end{document}